\documentclass{article}
\usepackage{ijcai19}
\usepackage[compact]{titlesec}   
\usepackage{adjustbox}
\usepackage{times}
\usepackage{soul}
\usepackage{booktabs}
\usepackage{url}
\usepackage[small]{caption}
\usepackage{graphicx}
\usepackage{amsmath}
\usepackage{booktabs}
\usepackage{algorithm}
\usepackage{algorithmic}
\usepackage{microtype}
\usepackage{graphicx}
\usepackage{booktabs} 
\usepackage{epsfig, psfrag, amstext,amsfonts,amssymb,subeqnarray,nccfoots,color,multirow}
\usepackage{amsthm}
\newtheorem{theorem}{Theorem}

\title{A General $\mathcal{O}(n^2)$ Hyper-Parameter Optimization for Gaussian Process Regression with Cross-Validation and Non-linearly Constrained ADMM}

\author{
    Linning Xu, Feng Yin, Jiawei Zhang, Zhi-Quan (Tom) Luo, Shuguang (Robert) Cui
    \affiliations
    School of Science and Engineering, The Chinese Univeristy of Hong Kong, Shenzhen \\
    Shenzhen Research Institute of Big Data (SRIBD), Guangdong Province, China 
    \emails
   \{linningxu, yinfeng, jiaweizhang2, luozq, shuguangcui\}@cuhk.edu.cn
}

\begin{document}

\maketitle

\begin{abstract}
Hyper-parameter optimization remains as the core issue of Gaussian process (GP) for machine learning nowadays. The benchmark method using maximum likelihood (ML) estimation and gradient descent (GD) is impractical for processing big data due to its $O(n^3)$ complexity. Many sophisticated global or local approximation models, for instance, sparse GP, distributed GP, have been proposed to address such complexity issue. In this paper, we propose two novel and general-purpose GP hyper-parameter training schemes (GPCV-ADMM) by replacing ML with cross-validation (CV) as the fitting criterion and replacing GD with a non-linearly constrained alternating direction method of multipliers (ADMM) as the optimization method. The proposed schemes are of $O(n^2)$ complexity for any covariance matrix without special structure. We conduct various experiments based on both synthetic and real data sets, wherein the proposed schemes show excellent performance in terms of convergence, hyper-parameter estimation accuracy, and computational time in comparison with the traditional ML based routines given in the GPML toolbox.  
\end{abstract}

\section{Introduction}
\label{sec:Introduction}
Gaussian process (GP) models constitute a class of important Bayesian non-parametric models for machine learning and are tightly connected to several other salient models, such as support vector machines (SVM), single-layer Bayesian neural networks, regularized-least-squares, relevance vector machines, auto-regressive-moving-average (ARMA) and deep neural networks \cite{Neal96,MacKay98,RW06,Mattews18}. The idea behind GP models is to impose a Gaussian prior on the underlying function and then compute the predictive distribution over the function given the observed data. Due to their outstanding performance in function approximation with a natural uncertainty bound, GP models have been adopted in a plethora of applications, e.g., nonlinear system identification \cite{Frigola13}, state-space modeling and trajectory prediction \cite{Frigola14}, financial data modeling and prediction \cite{Han16}, to mention a few. 

The predictive performance of GP regression models depends on the goodness of kernel selection and hyper-parameter estimation. There exist two classes of methods for GP hyper-parameter training. The first class of deterministic methods includes the maximum likelihood (ML) estimation based methods and the cross-validation (CV) based methods \cite{RW06,Krauth17} among the others. The second class of stochastic methods includes for instance, the hybrid Monte-Carlo and Markov chain Monte-Carlo sampling methods \cite{Neal97,Hensman15,Havasi18}. In this paper, we focus on the first class of deterministic methods.

The standard GP methods adopt the ML criterion to optimize the GP hyper-parameters via gradient descent (GD). However, the computational complexity scales cubically over the number of data points, i.e., $O(n^3)$, which is deemed as the most prominent weakness of such methods. Low-complexity GP methods could be obtained with the ideas to 1) find a smaller subset (with $m \ll n$ data points) of the complete data set and construct a sparse representation of the original covariance matrix, yielding $\mathcal{O}(m^2 n)$ complexity \cite{Candela05,Titsias09,Wilson15}; 2) adopt low-rank approximations of the covariance matrices, e.g., via hierarchical factorization of the covariance matrix into a product of block low-rank updates of the identity matrix, yielding $\mathcal{O}(n \log^{2} n)$ complexity \cite{Ambikasaran16}; 3) employ a number of $K$ local computing units, build GP models in smaller scales with a subset of data at each local computing unit, and fuse the hyper-parameter estimates via Bayesian committee machine type of approaches, yielding $\mathcal{O}(n^3/K^3)$ complexity \cite{Deisenroth15}. All these results rely on the use of ML criterion with different types of approximations. A more complete list of low-complexity methods were surveyed in \cite{Liu18b}. 

Cross-validation is yet another nice criterion for hyper-parameter training, demonstrated as an alternative of the ML criterion for model selection and GP hyper-parameter training in \cite{RW06}. The idea of using CV for GP hyper-parameter training is perhaps more desirable than using the ML criterion due to the following two reasons. First, for Gaussian noises, the CV based method is essentially equivalent to the ML based method. Second, for general noises, the CV based method directly minimizes the mean-squared-error (MSE), which is widely used as the ultimate performance metric for regression tasks. 

The main contributions of this paper are as follows. We propose two novel CV based schemes for GP hyper-parameter optimization. In order to eliminate the large-scale matrix inverse, an auxiliary variable is newly introduced and optimized jointly with the GP hyper-parameters using alternating direction method of multipliers (ADMM), where ADMM was first introduced in the mid-1970s and extended to handle a wide range of optimization problems in machine learning, control, and signal processing nowadays \cite{Eckstein92,Bertsekas97,Boyd11,Hong16}. The proposed schemes demonstrate $\mathcal{O}(n^2)$ computational complexity without making any approximation. Various experimental results show even better and faster GP hyper-parameter training results of the proposed schemes as compared to the GPML toolbox\footnote{Available on http://www.gaussianprocess.org/gpml/}.

The remainder of the paper is organized as follows. In Section~\ref{sec:Review}, we review the standard GP regression that applies ML and GD to train the hyper-parameters, yielding $\mathcal{O}(n^3)$ complexity. In Section~\ref{sec:HyperOpt-NewParadigm-General}, we introduce two new schemes based on CV and ADMM, whose $\mathcal{O}(n^2)$ complexity is validated in Section~\ref{sec:Theory}. Experimental results are given in Section~\ref{sec:Experiments}. Finally, we conclude the paper in Section~\ref{sec:Conclusion}. 

\section{Standard Gaussian Process Regression}
\label{sec:Review}
%
%
A Gaussian process is a collection of random variables, whose any finite subset follows a Gaussian distribution \cite{RW06}. In the sequel, we consider the scalar output, real valued Gaussian processes that are completely specified by a mean function and a kernel function (a.k.a. covariance function) as:
\begin{equation}
f(\boldsymbol{x}) \sim \mathcal{GP}(m(\boldsymbol{x}), k(\boldsymbol{x}, \boldsymbol{x}'; \boldsymbol{\theta}_{h})),
\end{equation} 
where $m(\boldsymbol{x}) $ denotes the mean function and $k(\boldsymbol{x}, \boldsymbol{x}'; \boldsymbol{\theta}_{h})$ represents the kernel function specified by a vector of hyper-parameters, $\boldsymbol{\theta}_{h}$.

The most widely used GP regression model is given by $y = f(\boldsymbol{x}) + e$, 
%
%
where $y \in \mathbb{R}$ is a continuous valued, scalar output. The unknown function $f(\boldsymbol{x}) : \mathbb{R}^{d} \mapsto \mathbb{R}$ is modeled via a zero mean Gaussian process for simplicity. The noise term $e$ is assumed to be Gaussian distributed with zero mean and variance $\sigma_{e}^{2}$, independent and identically distributed at different data points. The vector of unknown GP hyper-parameters is denoted by $\boldsymbol{\theta} \triangleq [\boldsymbol{\theta}_{h}^{T}, \sigma^{2}_{e}]^T$, whose dimension is assumed to be $p$.

In the training phase, we are given a data set $\mathcal{D} \triangleq \{\boldsymbol{X}, \boldsymbol{y} \}$, where $\boldsymbol{y} = [y_1, y_2, ..., y_n]^T$ is the set of outputs and $\boldsymbol{X}=[\boldsymbol{x}_1, \boldsymbol{x}_2,...,\boldsymbol{x}_n]$ is the set of inputs. As the benchmark, we review the classic maximum likelihood  (ML) based GP hyper-parameter optimization. Due to the Gaussian assumption on the noise term, the log-likelihood function could be obtained in closed form, and the GP hyper-parameters could be trained equivalently by minimizing the negative log-marginal likelihood function (ignoring the unrelated terms) as: 
\begin{equation}
\boldsymbol{\theta}_{ML} \triangleq \! \arg \min_{\boldsymbol{\theta}} \,  l(\boldsymbol{\theta}) \!=\! \boldsymbol{y}^T \boldsymbol{C}^{-1}(\boldsymbol{\theta}) \boldsymbol{y} \!+\!  \log \det \left( \boldsymbol{C}(\boldsymbol{\theta}) \right), 
\label{eq:likelihood}
\end{equation}
where $\boldsymbol{C}(\boldsymbol{\theta}) \triangleq \boldsymbol{K}(\boldsymbol{X}, \boldsymbol{X}; \boldsymbol{\theta}_h) + \sigma_{e}^{2} \boldsymbol{I}_{n}$. 
The GD type of methods are most widely used for hyper-parameter optimization in the GP community. At the $(k+1)$-th iteration, each element of the GP hyper-parameters is updated as:
\begin{equation}
\theta_{i}^{k+1} = \theta_{i}^{k} - \mu \cdot \frac{\partial l(\boldsymbol{\theta}) }{\partial \theta_i} |_{\boldsymbol{\theta} = \boldsymbol{\theta}^{k}},  \quad \forall i = 1,2,...,p,
\end{equation}
where $\mu$ is a positive step size. The partial derivative of the $i$-th element in $\boldsymbol{\theta}$ can be derived in closed form as:
\begin{equation}
\frac{\partial l(\boldsymbol{\theta}) }{\partial \theta_i} = \textrm{tr} \left\lbrace \left( \boldsymbol{C}^{-1}(\boldsymbol{\theta}) - \boldsymbol{\gamma} \boldsymbol{\gamma}^{T} \right) \frac{\partial \boldsymbol{C}(\boldsymbol{\theta})}{\partial \theta_i} \right\rbrace,
\end{equation}
where $\boldsymbol{\gamma} \triangleq \boldsymbol{C}^{-1}(\boldsymbol{\theta}) \boldsymbol{y}$ is defined for notational brevity. At each iteration, $\boldsymbol{C}^{-1}(\boldsymbol{\theta})$ needs to be re-evaluated with the updated $\boldsymbol{\theta} = \boldsymbol{\theta}^{k}$, where multiplication of $n \times n$ matrices has to be performed for several times. Therefore, the total computational complexity scales as $\mathcal{O}(n^3)$ per iteration, in general. 

In the test phase, the trained GP models compute the predictive distribution of $\boldsymbol{y}_{*} =  [y_{*,1}, y_{*,2},...,y_{*,n_{*}}]^T$ for novel test inputs $\boldsymbol{X}_{*} = [\boldsymbol{x}_{*,1}, \boldsymbol{x}_{*,2},...,\boldsymbol{x}_{*,n_{*}}]$. The test data set is denoted as $\mathcal{D}_{*} = \{ \boldsymbol{y}_{*}, \boldsymbol{X}_{*}\}$. 
%
%
According to the definition of Gaussian process and applying the canonical results of conditional Gaussian distribution, we could easily derive the predictive distribution to be $p(\boldsymbol{y}_{*} \vert \mathcal{D}, \boldsymbol{X}_{*}; \boldsymbol{\theta}) \sim
\mathcal{N} \left(  \bar{\boldsymbol{m}} , \bar{\boldsymbol{V}}  \right)$, 
%
%
where the mean and variance are:
\begin{align}
\bar{\boldsymbol{m}} & = \boldsymbol{K}(\boldsymbol{X}_{*}, \boldsymbol{X}) \left[ \boldsymbol{K}(\boldsymbol{X}, \boldsymbol{X}) + \sigma_{e}^{2} \boldsymbol{I}_{n} \right]^{-1} \boldsymbol{y}, \label{eq:posterior-mean} \\
\bar{\boldsymbol{V}} & = \boldsymbol{K}(\boldsymbol{X}_{*}, \boldsymbol{X}_{*}) + \sigma_{e}^{2} \boldsymbol{I}_{n_{*}} \nonumber \\ 
&- \boldsymbol{K}(\boldsymbol{X}_{*}, \boldsymbol{X}) \left[ \boldsymbol{K}(\boldsymbol{X}, \boldsymbol{X}) + \sigma_{e}^{2} \boldsymbol{I}_{n} \right]^{-1}  \boldsymbol{K}(\boldsymbol{X}, \boldsymbol{X}_{*}). \label{eq:posterior-var}
\end{align}
Here, $\boldsymbol{K}(\boldsymbol{X}, \boldsymbol{X})$ is an $n \times n$ matrix of covariances among the training inputs; $\boldsymbol{K}(\boldsymbol{X}, \boldsymbol{X}_{*})$ is an $n \times n_{*}$ matrix of covariances between the training inputs and test inputs; $\boldsymbol{K}(\boldsymbol{X}_{*}, \boldsymbol{X}_{*})$ is an $n_{*} \times n_{*}$ matrix of covariances among the test inputs.

\section{Proposed Cross-Validation based GP Hyper-parameter Optimization Schemes}
\label{sec:HyperOpt-NewParadigm-General}
%
In this section, we introduce two new GP hyper-parameter optimization schemes by replacing ML with CV, and replacing GD with ADMM. The proposed schemes achieve $O(n^2)$ computational complexity. We assume that the number of training data samples, $n$, is moderately large such that a single computing unit could handle computations in the order of $\mathcal{O}(n^2)$, while $\mathcal{O}(n^3)$ is beyond the processing limit.  
%

\subsection{Hold-out Cross-Validation based Scheme} 
\label{sec:Scheme-I}
We first focus on the training phase and consider the simple hold-out cross-validation (HOCV), which is widely used for model selection in machine learning. We divide the data set $\mathcal{D}$ into two non-overlapping subsets, namely the training set  $\mathcal{D}_{T} =  \{\boldsymbol{X}_{T},\boldsymbol{y}_{T} \}$ with $|\mathcal{D}_{T} | = n_t$ samples and the validation set $\mathcal{D}_{V} = \{\boldsymbol{X}_{V},\boldsymbol{y}_{V}\}$ with $|\mathcal{D}_{V} | = n_v$ samples. Note that, the ML based scheme instead use the whole $\mathcal{D}$ for training the GP hyper-parameters. We illustrate this difference with an example in Figure~\ref{fig:figure1}. In practice, we could pick $n_v$ samples randomly from $\mathcal{D}$ for validation and leave the rest for training. 
The overview of Gaussian process regression in Section~\ref{sec:Review} tells us that the predictive mean of the validation points in $\mathcal{D}_{V}$ given the training data set $\mathcal{D}_{T}$ is: 
\begin{align}
\bar{\boldsymbol{m}}(\boldsymbol{X}_{V}; \boldsymbol{\theta}) \triangleq \boldsymbol{K}(\boldsymbol{X}_V, \boldsymbol{X}_T;\boldsymbol{\theta}_h)\boldsymbol{z}_T,
\end{align}
where 
\begin{equation}
\boldsymbol{z}_T \triangleq 
\left[ \boldsymbol{K}(\boldsymbol{X}_T, \boldsymbol{X}_T; \boldsymbol{\theta}_h) + \sigma_{e}^{2} \boldsymbol{I}_{n} \right]^{-1}\boldsymbol{y}_{T}.
\end{equation}
%
In the sequel, we use the short forms $\boldsymbol{K}_{VT}(\boldsymbol{\theta}_h)$, $\boldsymbol{K}_{TT}(\boldsymbol{\theta}_h)$ and $\boldsymbol{C}(\boldsymbol{\theta})$ to denote $\boldsymbol{K}(\boldsymbol{X}_V, \boldsymbol{X}_T;\boldsymbol{\theta}_h)$, $\boldsymbol{K}(\boldsymbol{X}_T, \boldsymbol{X}_T;\boldsymbol{\theta}_h)$, and $\boldsymbol{K}(\boldsymbol{X}_T, \boldsymbol{X}_T;\boldsymbol{\theta}_h) + \sigma_{e}^{2} \boldsymbol{I}_{n}$ respectively.

We aim to find an optimal vector of the GP hyper-parameters that minimizes the differences between the validation outputs and their predictive means, which is formulated as:
\begin{equation}
\boldsymbol{\theta}_{CV} = \arg \min_{\boldsymbol{\theta}} \left| \left| \boldsymbol{y}_V - \bar{\boldsymbol{m}}(\boldsymbol{X}_{V}; \boldsymbol{\theta})\ \right| \right|_{2}^{2}.
\label{eq:new-objective-func}
\end{equation}
This optimization problem is non-convex in terms of $\boldsymbol{\theta}$ for most kernels. When applying GD for solving (\ref{eq:new-objective-func}), the inverse of a possibly large $n \times n$ covariance matrix $\boldsymbol{C}(\boldsymbol{\theta})$ has to be computed at each iteration with $\mathcal{O}(n^3)$ complexity in general, which forbids the practical use of Gaussian process regression for big data applications. In order to remedy this drawback, we let $\boldsymbol{z}_T$ be a vector-formed auxiliary variable and impose the following nonlinear equality constraint: 
 \begin{equation}
\boldsymbol{C}(\boldsymbol{\theta}) \boldsymbol{z}_T = \boldsymbol{y}_{T}.
\label{eq:equality-constraint}
\end{equation}
In order to adopt ADMM for GP hyper-parameter optimization with $\mathcal{O}(n^2)$ complexity, we formulate the augmented Lagrangian function as:
\begin{align}
L_{\rho} \left( \boldsymbol{\theta}, \boldsymbol{z}_T, \boldsymbol{\lambda} \right) &\triangleq \left| \left| \boldsymbol{y}_V  -  \boldsymbol{K}_{VT}(\boldsymbol{\theta}_h) \boldsymbol{z}_T  \right| \right|_{2}^{2} \nonumber \\ 
& + \! \boldsymbol{\lambda}^{T} \! \left( \boldsymbol{C}(\boldsymbol{\theta}) \boldsymbol{z}_T \!-\! \boldsymbol{y}_{T} \right) +  \frac{\rho}{2}\left|\left| \boldsymbol{C}(\boldsymbol{\theta}) \boldsymbol{z}_T \!-\! \boldsymbol{y}_{T} \right|\right|_{2}^{2}, \nonumber
\end{align}
where the regularization parameter $\rho \geq 0$ is pre-selected and $\boldsymbol{\lambda}$ is a vector of Lagrange multipliers. The complete method consists of a $\boldsymbol{\theta}$-minimization step, a $\boldsymbol{z}_T$-minimization step, and a closed form dual variable update step. Concretely, at the $(\eta+1)$-th iteration we have,
\begin{subequations}
\begin{align}
\boldsymbol{\theta}^{\eta+1} &=  \arg\min_{\boldsymbol{\theta}} 
L_{\rho}(\boldsymbol{\theta}, \boldsymbol{z}_T^{\eta}, \boldsymbol{\lambda}^{\eta}), \label{eq:thetaUpd}\\
\boldsymbol{z}_T^{\eta+1} &=  \arg\min_{\boldsymbol{z}_T} 
L_{\rho}(\boldsymbol{\theta}^{\eta+1}, \boldsymbol{z}_T, \boldsymbol{\lambda}^{\eta}), \label{eq:zUpd}\\
\boldsymbol{\lambda}^{\eta+1} &=  \boldsymbol{\lambda}^{\eta} +
\rho \left[ \boldsymbol{C}(\boldsymbol{\theta}^{\eta+1}) \boldsymbol{z}_T^{\eta+1} - \boldsymbol{y}_{T} \right].
\label{eq:lambdaUpd}
\end{align}
\end{subequations}

We elaborate on the $\boldsymbol{\theta}$-minimization step first. Note that $L_{\rho}(\boldsymbol{\theta}, \boldsymbol{z}_T^{\eta}, \boldsymbol{\lambda}^{\eta})$ is often a non-convex function in terms of $\boldsymbol{\theta}$, which could be minimized as:
\begin{align}
\boldsymbol{\theta}^{\eta+1} &= \boldsymbol{\theta}^{\eta} -
\mu_{1} \cdot  \nabla_{\boldsymbol{\theta}} L_{\rho}(\boldsymbol{\theta}, \boldsymbol{z}_T^{\eta}, \boldsymbol{\lambda}^{\eta})  \nonumber \\
&\equiv  \boldsymbol{\theta}^{\eta} - \mu_{1} \cdot \nabla_{\boldsymbol{\theta}} \boldsymbol{g}^{(\eta)}\left(\boldsymbol{\theta} \right)\mid_{\boldsymbol{\theta} = \boldsymbol{\theta}^{\eta}},
\label{eq:gradient-descent for theta}
\end{align}
where $\mu_{1}$ is a positive step size, $\boldsymbol{g}^{(\eta)}\left(\boldsymbol{\theta} \right)$ and its partial derivatives are introduced in the Appendix. When $\mu_1$ is chosen via line minimization or the Armijo's rule, the sequence $\{\boldsymbol{\theta}^{\eta} \}$ is guaranteed to converge to a stationary point of subproblem (\ref{eq:thetaUpd}) according to Proposition 1.2.1 of \cite{Bertsekas16}.
%
%

Second, we elaborate on the $\boldsymbol{z}_T$-minimization step. In subproblem (\ref{eq:zUpd}), it is easy to verify that $L_{\rho}(\boldsymbol{\theta}^{\eta+1}, \boldsymbol{z}_T, \boldsymbol{\lambda}^{\eta})$ is a quadratic function of $\boldsymbol{z}_T$ and minimizing it with respect to $\boldsymbol{z}_T$ is equivalent to
\begin{equation}
\arg \min_{\boldsymbol{z}_T} \boldsymbol{g}^{(\eta)}\left( \boldsymbol{z}_T \right) = (\boldsymbol{b}^{\eta})^{T} \boldsymbol{z}_{T} + \boldsymbol{z}_{T}^{T} \boldsymbol{\Sigma}^{\eta} \boldsymbol{z}_{T},
\label{eq:cost-func-zt}
\end{equation} 
where 
\begin{align}
\boldsymbol{b}^{\eta} &\triangleq
\boldsymbol{C}(\boldsymbol{\theta}^{\eta+1}) \boldsymbol{\lambda} -  \rho \boldsymbol{C}(\boldsymbol{\theta}^{\eta+1})\boldsymbol{y}_{T} -2  \boldsymbol{K}_{VT}^{T}(\boldsymbol{\theta}_h^{\eta+1}) \boldsymbol{y}_{V}, \nonumber \\
\boldsymbol{\Sigma}^{\eta} &\triangleq 
\boldsymbol{K}_{VT}^{T}(\boldsymbol{\theta}_h^{\eta+1})  \boldsymbol{K}_{VT}(\boldsymbol{\theta}_h^{\eta+1}) +  \frac{\rho}{2} \boldsymbol{C}^{2}(\boldsymbol{\theta}^{\eta+1}) .  \nonumber 
\end{align}
It is easy to verify that $\boldsymbol{\Sigma}^{\eta}$ is always positive definite. However, taking the derivative of $\boldsymbol{g}^{(\eta)}\left( \boldsymbol{z}_T \right)$  with respect to $\boldsymbol{z}_T$ and setting it equal to zero for closed form solution involves the inverse of $\boldsymbol{\Sigma}^{\eta}$ with $\mathcal{O}(n^3)$ complexity. Thus, we instead solve the quadratic minimization problem in (\ref{eq:cost-func-zt}) numerically via the conjugate gradient method (CGM) \cite{Bertsekas16}. Specifically, we update the auxiliary variable as
\begin{equation}
\boldsymbol{z}_{T}^{\eta+1} = \boldsymbol{z}_{T}^{\eta} + \mu_2 \cdot \boldsymbol{d}^{\eta}_{z},
\label{eq:CGM-update-zt}
\end{equation} 
where $\mu_2$ is a stepsize obtained via line minimization, the descent direction $\boldsymbol{d}^{\eta}_{z}$ is an outcome of Gram-Schmidt orthogonization process: $\boldsymbol{d}^{\eta}_{z} = -\boldsymbol{g}^{\eta}_{z} + \beta_r \boldsymbol{d}^{\eta-1}_{z}$
%
%
with $\boldsymbol{g}^{\eta}_{z}$ being short for $\nabla_{\boldsymbol{z}_T} g(\boldsymbol{z}^{\eta}_T) = 2 \boldsymbol{\Sigma}^{\eta} \boldsymbol{z}^{\eta}_T + \boldsymbol{b}^{\eta}$ and $\beta_r = || \boldsymbol{g}^{\eta}_{z} ||_{2}^{2}/ || \boldsymbol{g}^{\eta-1}_{z} ||_{2}^{2}$
%
%
, which starts from $\boldsymbol{d}^{0}_{z} = -\boldsymbol{g}^{0}_{z} = -\nabla_{\boldsymbol{z}_T} g(\boldsymbol{z}^{0}_T)$ and terminates at an optimal solution after at most $n_T$ steps. Recall that we introduce an auxiliary variable $\boldsymbol{z}_T$ to eliminate the large matrix inverses, fortunately the efficient CGM exists for the solution with little computational effort. 


Lastly, the update of $\boldsymbol{\lambda}^{\eta+1}$ is conducted in light of (\ref{eq:lambdaUpd}) after $\boldsymbol{\theta}^{\eta+1}$ and $\boldsymbol{z}_T^{\eta+1}$ are obtained. For the HOCV based scheme, we have the following theorem (for more details see Sections 4.2 and 5.2 of \cite{Bertsekas16}):
\begin{theorem}
Local Convergence Property: When taking the initial guess $\boldsymbol{\lambda}^{0}$ close to the optimal Lagrange multiplier $\boldsymbol{\lambda}^{*}$ and taking $\rho$ large enough, solving the unconstrained minimization problem $L_{\rho}(\boldsymbol{\theta}, \boldsymbol{z}_{T}, \boldsymbol{\lambda})$ can yield points close to the local minimum $\boldsymbol{\theta}^{*}$ and $\boldsymbol{z}_{T}^{*}$ that satisfy the sufficient optimality conditions. 
\end{theorem}

\subsection{$K$-fold Cross Validation Based Scheme}
\label{sec:Scheme-II}
%

In this subsection, we aim to design a $K$-fold CV based GP hyper-parameter optimization scheme, which is able to generate more robust result and exploit parallel computing. We let $\mathcal{D}_{V}^{(k)} = \{\boldsymbol{y}_{V}^{(k)},  \boldsymbol{X}_{V}^{(k)}\}$ be the $k$-th partition of the complete data set $\mathcal{D}$ to be used for validation, and the corresponding training set $\mathcal{D}_{T}^{(k)} = \mathcal{D} \backslash \mathcal{D}_{V}^{(k)}$. 

\paragraph{A Naive Scheme:} We train the GP hyper-parameters $\boldsymbol{\theta}_{CV}^{(k)}$, heuristically for every partition $k=1,2,...,K$, using the same routine given in the HOCV based scheme (see Algorithm~\ref{alg:algorithm1}), and average the results to obtain the final estimate $\boldsymbol{\theta}_{CV} = 1/K \cdot \sum_{k=1}^{K} \boldsymbol{\theta}_{CV}^{(k)}$. 

\paragraph{A Principled Scheme:} Alternatively, we formulate an optimization problem for the same purpose but with a sound rationale. For $K \geq 2$, the final estimate of the GP hyper-parameters is obtained as:
\begin{equation}
\boldsymbol{\theta}_{CV} = \arg \min_{\boldsymbol{\theta}} \sum_{k=1}^{K} \left| \left| \boldsymbol{y}_{V}^{(k)} - \bar{\boldsymbol{m}}(\boldsymbol{X}_{V}^{(k)}; \boldsymbol{\theta})\ \right| \right|_{2}^{2}.
\label{eq:new-objective-func3}
\end{equation}
%
%
%
To tackle (\ref{eq:new-objective-func3}), we introduce some local copies of $\boldsymbol{\theta}$ and solve the following linear equality-constrained optimization problem:
\begin{align}
\boldsymbol{\theta}_{CV} &= \arg \min_{\boldsymbol{\theta}_1,...,\boldsymbol{\theta}_K } \sum_{k=1}^{K}  l_{k}(\boldsymbol{\theta}_k) \nonumber \\
\textrm{s.t.} & \quad \boldsymbol{\theta}_1 = \boldsymbol{\theta}_2 =,...,= \boldsymbol{\theta}_K = \boldsymbol{z},
\label{eq:new-objective-func2}
\end{align}
where $l_{k}(\boldsymbol{\theta}_k) \triangleq \left| \left| \boldsymbol{y}_{V}^{(k)} - \bar{\boldsymbol{m}}(\boldsymbol{X}_{V}^{(k)}; \boldsymbol{\theta}_k)\ \right| \right|_{2}^{2}$ is non-convex in terms of $\boldsymbol{\theta}_k$ for most kernels. The following theorem from \cite{Hong16} is valuable and supportive:
\begin{theorem}
When the following assumptions hold: 1) $l_{k}(\boldsymbol{\theta}_k)$ satisfies the Lipschitz condition; 2) the augmented Lagrangian parameter, $\rho$, is chosen large enough; 3) the minimization problems are bounded from below and all ADMM subproblems are solved exactly, etc., it is guaranteed that any limit point of problem (\ref{eq:new-objective-func2}) is also a stationary solution.
\end{theorem}
\begin{algorithm}[t]
\caption{HOCV Based GP Hyper-Parameter Optimization}
\textbf{Input}: Complete data set $\mathcal{D}$ divided into $\mathcal{D}_T$ and $\mathcal{D}_V$ \\
\textbf{Output}: Optimal GP hyper-parameters $\boldsymbol{\theta}^{*}$ \\
\textbf{Initialization}: $\eta=0$, $\boldsymbol{\lambda}^{0}$, $\boldsymbol{z}_{T}^{0}$, $\boldsymbol{\theta}^{0}$
\begin{algorithmic}[1]
\WHILE{$|| \boldsymbol{\theta}^{\eta+1} - \boldsymbol{\theta}^{\eta} ||_{2} \geq \epsilon $ and $\eta \leq maxItr$}
\STATE Update $\boldsymbol{\theta}^{\eta+1}$ according to (\ref{eq:gradient-descent for theta})		
\STATE Update $\boldsymbol{z}_{T}^{\eta+1}$ according to (\ref{eq:CGM-update-zt})
\STATE Update $\boldsymbol{\lambda}^{\eta+1}$ according to (\ref{eq:lambdaUpd})
\STATE Set $\eta = \eta +1$.
\ENDWHILE
\STATE \textbf{return} $\boldsymbol{\theta}^{*} = \boldsymbol{\theta}^{\eta}$
\end{algorithmic}
\label{alg:algorithm1}
\end{algorithm}

%

\section{Computational Complexity}
\label{sec:Theory}
%
%

We aim to verify the computational complexity of the proposed GP hyper-parameter optimization schemes in Section~\ref{sec:HyperOpt-NewParadigm-General}. 

We start with the HOCV based scheme. 
Updating one particular element of the hyper-parameters (out of $p$ elements), say $\theta_i$, according to (\ref{eq:gradient-descent for theta}), mainly involves the computations of $\frac{\partial \boldsymbol{K}_{VT}(\boldsymbol{\theta}_h)}{\partial \theta_i}  \boldsymbol{z}_T^{\eta}$, and $\boldsymbol{K}_{VT}(\boldsymbol{\theta}_h) \boldsymbol{z}_T^{\eta}$, $\frac{\partial \boldsymbol{K}_{TT}(\boldsymbol{\theta}_h)}{\partial \theta_i} \boldsymbol{z}_T^{\eta}$, $\boldsymbol{K}_{TT}(\boldsymbol{\theta}_h)  \boldsymbol{z}_T^{\eta}$ and some cheap vector inner products. Therefore, the computational complexity scales as $\mathcal{O}(n_v \cdot n_t + n_t^2) = \mathcal{O}(n \cdot n_t)$ for this step. Similarly, updating the auxiliary parameter $\boldsymbol{z}_T$, according to (\ref{eq:CGM-update-zt}), mainly involves the computations of $\boldsymbol{K}_{VT}(\boldsymbol{\theta}_h^{\eta+1}) \boldsymbol{z}_T^{\eta}$, $\boldsymbol{C}(\boldsymbol{\theta}^{\eta+1}) \boldsymbol{z}_T^{\eta}$ as well as some cheap vector inner products (for details see the Appendix), thus the computational complexity also scales as $\mathcal{O}(n_v \cdot n_t + n_t^2) = \mathcal{O}(n \cdot n_t)$. The third step involves only a closed form update, whose complexity scales as $\mathcal{O}(n_t^2)$. Therefore, the overall computational complexity for running one complete ADMM iteration scales as $\mathcal{O}(p \cdot n \cdot n_t) \approx \mathcal{O}(n^2)$ for $p \ll n$, which is much lower than $\mathcal{O}(n^3)$.

For the $K$-fold CV based scheme, each computing unit updates a local copy of the global variable, $\boldsymbol{\theta}_k$, incurring $\mathcal{O}(n^2)$ complexity according to the above analysis. The overall complexity remains low for practical $K$ and $p$ values. 


\section{Experiments}
\label{sec:Experiments}
In this section, we aim to evaluate our proposed schemes, termed as GPCV-ADMM, using both synthetic and real data sets. Specifically, we choose the naive two-fold CV based scheme, which is more practical to implement. As the benchmark, we choose the state-of-the-art ML based scheme given in the GPML toolbox.

\begin{figure}[t]
\centering
\includegraphics[width=\linewidth]{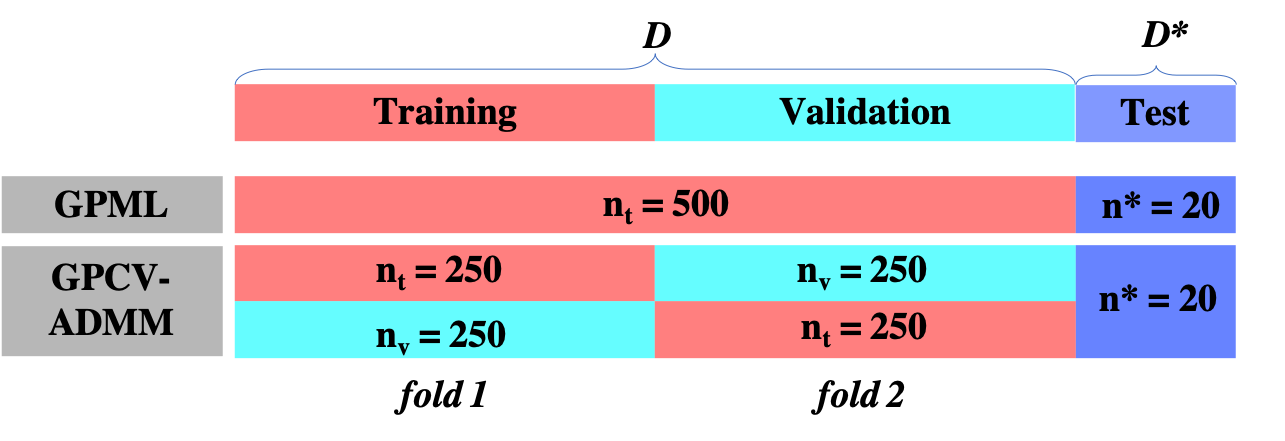} \\
\caption{Illustration of the different data compositions used bt GPML and GPCV-ADMM schemes(with sample size $n=500$). In the training phase, GPML treats the whole $\mathcal{D}$ as the training data set, while GPCV-ADMM separates $\mathcal{D}$ into two equal-sized partitions for cross-validation purpose. In the test phase, another data set $\mathcal{D}_{*}$ with $n^* = 20$ samples is used for evaluating their prediction performance.}
\label{fig:figure1}
\end{figure}

\paragraph{Simulation Platform:} Our GPCV-ADMM is implemented in R (version 3.5.2), and compared with the GPML toolbox executed in MATLAB 2018b. All the experiments were conducted on a MacBook Pro with 2.2 GHz Intel Core i7.

\paragraph{Algorithmic Setup:} For both fairness and clarity of comparisons, all the hyper-parameters are initialized with same values for both GPCV-ADMM and GPML when testing on the synthetic data sets under a fixed kernel configuration. However, random restarts are recommended for initialization in practice, and was also adopted in our experiment for the real $CO_2$ concentration data set. The auxiliary variable $\boldsymbol{z}_T$ of GPCV-ADMM is initialized according to (\ref{eq:equality-constraint}) with a perturbed $\boldsymbol{\theta}^0$, and the dual variable $\boldsymbol{\lambda}$ initialized to be a vector of all ones. The regularization parameter is pre-selected to be $\rho=5$. The error tolerance for ADMM is set to be $\epsilon = 10^{-2}$ and the maximum number of iterations is set to be $maxItr = 100$. The source code is available online\footnote{  Available on https://github.com/eveneveno/GPCV-ADMM-2019}.

\subsection{Data Sets}
\paragraph{Synthetic Data:} The synthetic data sets used in our experiments were generated from: 1) a squared exponential (SE) kernel with $\boldsymbol{\theta} = l$, 2) a local periodic (LP) kernel with $\boldsymbol{\theta} = [l, p]^T$, and 3) a composite kernel by adding up a SE and LP kernel (short as SE+LP in the sequel) with $\boldsymbol{\theta} = [l_1, l_2, p]^T$. Explicit forms of these selected kernels and their interpretations are given in the Appendix. For each selected kernel, we generated three kinds of data sets with sample sizes $n = 500, 1000, 2000$, respectively, for the primary purpose of verifying the $O(n^2)$ complexity of GPCV-ADMM as compared to the $O(n^3)$ complexity of GPML. The synthetic data inputs lie in the range of $[0, 10]$ for $n=500$, while in the range of $[0, 20]$ for $n=1000, 2000$. For a fixed kernel configuration and a fixed sample size, we ran 50 independent Monte Carlo trials to evaluate the goodness of the hyper-parameter estimates obtained by the two schemes. Note that the two schemes use the full data sets differently for training the GP hyper-parameters, as is shown in Figure~\ref{fig:figure1} for clarity.\\  

\paragraph{Real Atmospheric $CO_2$ Concentration Data:} This data set consists of the monthly average atmospheric $CO_2$ concentrations at the Mauna Loa Observatory, Hawaii, from 1958 to 2015, and is widely used in the GP community\footnote{Available on http://scrippsco2.ucsd.edu/data/atmospheric\_co2}. In \cite{RW06}, the authors proposed to use a composite kernel that consists of four parts: 1) an SE kernel modeling the rising trend, 2) an LP kernel representing the seasonal repetition, 3) a rational quadratic kernel for small irregularities, and  4) and another SE kernel for the noise term. Here, we adopt an SE+LP kernel to capture the main features of the data. The explicit expression of the adopted kernel is given in the Appendix. 

\subsection{Result Analysis}
\begin{table*}[t]
\resizebox{2.05\columnwidth}{!}{%
\begin{tabular}{lllrrrr} \toprule
 & & & \multicolumn{2}{c}{Hyper-paramete Estimates (\textcolor{red}{std})} &  \multicolumn{2}{c}{Test MSE} \\
Kernel	& Setting &Size & GPCV-ADMM  &GPML & GPCV-ADMM  &GPML \\ \hline

SE& \textbf{$l=0.5$} &500     &   [0.46(\textcolor{red}{0.054})]     &   [0.52(\textcolor{red}{0.018})]     &  \textbf{0.12} &  0.13    \\
 &&1000    &  [0.50(\textcolor{red}{0.044})]      & [0.50(\textcolor{red}{0.017})]  &  \textbf{0.12}   &   0.14 \\ 
&&2000    & [0.50(\textcolor{red}{0.017})]      &   [0.53(\textcolor{red}{0.020})]   &  \textbf{0.12}  & 0.14  \\ 

LP& \textbf{$l=0.5$} & 500   &     [0.34(\textcolor{red}{0.023}),1.13(\textcolor{red}{0.080})]       &    [0.55(\textcolor{red}{0.052}),1.06(\textcolor{red}{0.090})]     &   \textbf{0.13}  &     0.36     \\ 
&\textbf{$p=1$}& 1000    & [0.39(\textcolor{red}{0.014}),1.06(\textcolor{red}{0.063})]  & [0.52(\textcolor{red}{0.058}),1.16(\textcolor{red}{0.130})]  &  \textbf{0.17} &    0.44  \\
&&2000 &    [0.44(\textcolor{red}{0.082}),1.19(\textcolor{red}{0.008})]     & [0.53(\textcolor{red}{0.013}),1.02(\textcolor{red}{0.019})]  &   \textbf{0.26} &     0.28   \\ 

SE+LP  & \textbf{$l_1=3$} &500  &    [3.73(\textcolor{red}{0.280}),0.94(\textcolor{red}{0.100}),2.38(\textcolor{red}{0.150})]   &  [3.62(\textcolor{red}{0.630}), 1.08(\textcolor{red}{0.210}),2.38(\textcolor{red}{0.860})]    &  \textbf{0.18}  & 0.21 \\ 
&\textbf{$l_2=1$}& 1000&  [3.74(\textcolor{red}{0.250}),0.95(\textcolor{red}{0.110}),2.14(\textcolor{red}{0.090})]     & [3.61(\textcolor{red}{0.560}),1.05(\textcolor{red}{0.180}),2.27(\textcolor{red}{0.600})]    &  \textbf{0.13}    &  0.15 \\
&\textbf{$l_3=2$}&2000 &  [3.94(\textcolor{red}{0.075}),0.99(\textcolor{red}{0.120}),2.10(\textcolor{red}{0.080})]      & [3.69(\textcolor{red}{0.390}),1.09(\textcolor{red}{0.130}),2.13(\textcolor{red}{0.290})]      &  \textbf{0.34}  &     0.37 \\ \bottomrule
\end{tabular}%
}
\caption{Quantitative comparisons between GPCV-ADMM and GPML across nine synthetic data sets (combining three kernels and three data lengths). We recorded 1) the sample mean of hyper-parameter estimates and 2) the sample mean of the MSE averaged over 50 Monte Carlo simulations. }
%
\label{tab:table1}
\end{table*}

\paragraph{Conclusion:} The following experimental results confirm that GPCV-ADMM is able to achieve comparable (even better) hyper-parameter estimation performance compared to GPML with much reduced computation complexity. 

\paragraph{Estimation Performance:}
Table~\ref{tab:table1} gives the quantitative comparisons between GPCV-ADMM and GPML across all synthetic data sets, where the underlying true hyper-parameter values are given as references. The results show that GPCV-ADMM hyper-parameter estimates are fairly close to the GPML estimates, and moreover they are both close to the true values.
It is not surprising that GPCV-ADMM tends to generate lower test MSE ($MSE_{test}=  \frac{1}{N} \left| \left| \boldsymbol{y}_* - \bar{\boldsymbol{m}}(\boldsymbol{X}_*; \boldsymbol{\theta}^{*})\ \right| \right|_{2}^{2}$), simply because GPCV-ADMM is designed intentionally to minimize MSE and thus better reveals the predictive performance compared with the ML based scheme. According to the Monte Carlo simulation results, GPCV-ADMM estimation is more robust, demonstrating smaller sample standard deviation of both the hyper-parameter estimates and the test MSE.

In the above tests with synthetic data sets, we implicitly assumed that the underlying kernel function was precisely known, i.e., there was no model mismatch. In the following test performed on the real atmospheric $CO_2$ data, recorded from 1958 to 2015, GPCV-ADMM demonstrated outstanding training and prediction performance as shown in Figure~\ref{fig:figure4}, despite of model mismatch. We use the $CO_2$ data ranging from 1958 to 2008 for training. The hyper-parameter estimates obtained by the GPML in \cite{RW06} with several random restarts are $l_1 = 67$ years, $l_2 = 90$ years, and $l_3 = 1.3$ respectively. Though the adopted kernel function in our experiment is only a portion of the original one in \cite{RW06}, it could be viewed as a more general function approximation with the ignorance of some small irregularities and noises, thus these reference values are still considered to be good enough. For GPCV-ADMM, we tried a few random restarts and picked the one with the lowest training MSE as suggested in  \cite{RW06}. We obtained $l_1 = 27$ years, $l_2 = 51$ years, and $l_3 = 1.26$. 
To evaluate the predictive performance of the two competing schemes, we predicted the real $CO_2$ concentrations from 2009 to 2015. The standardized test MSE obtained by GPML is 1.408, while our GPCV-ADMM gives a better prediction with a lower test MSE of 1.307. 

\begin{figure}[h]
\centering
\includegraphics[width=\linewidth]{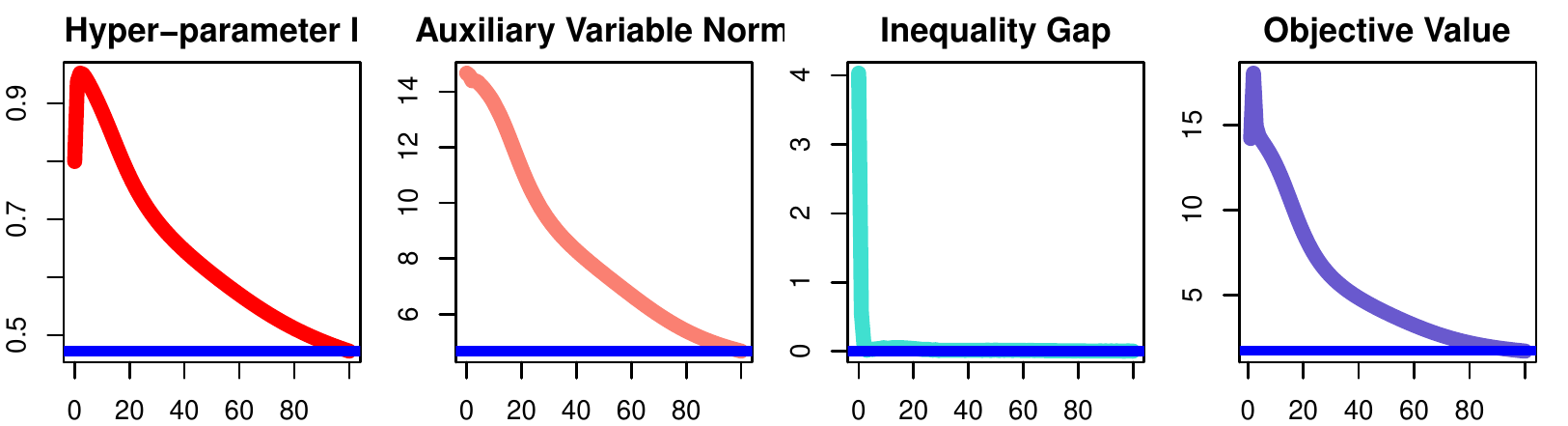} \\
\includegraphics[width=\linewidth]{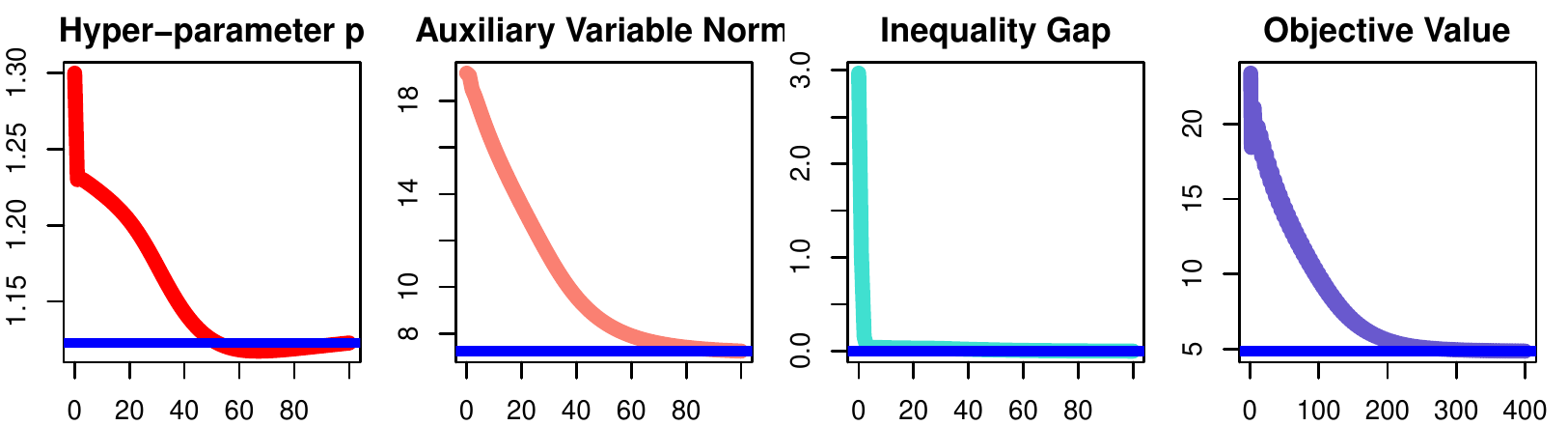} \\
\includegraphics[width=\linewidth]{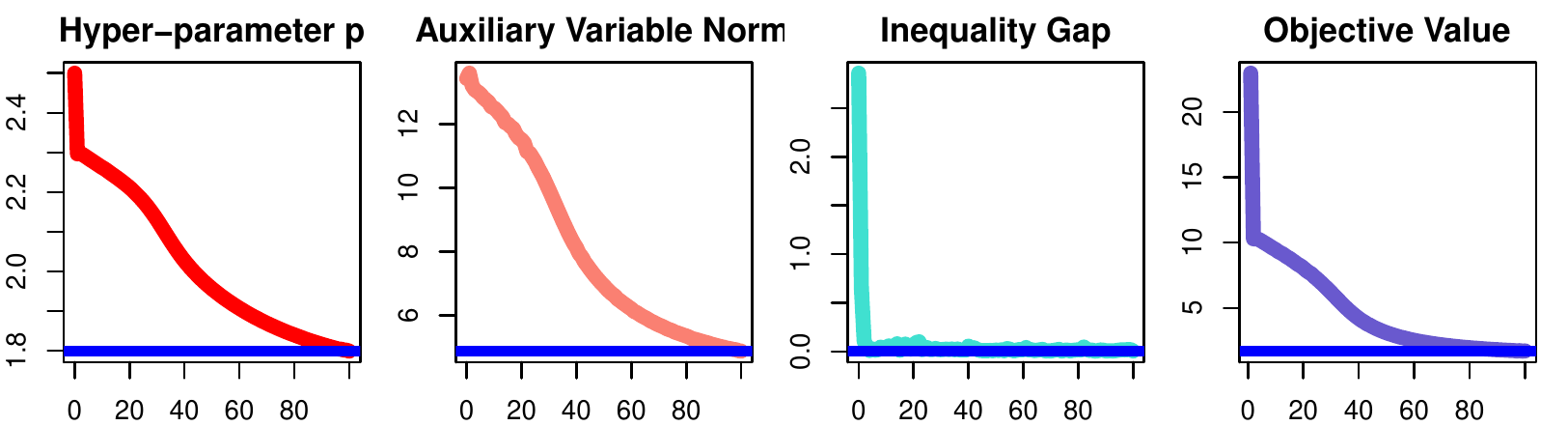} \\
\caption{Convergence curves of 1) a representative GP hyper-parameter, 
2) the $L_2$ norm of the auxiliary variable, $||\boldsymbol{z}_T||_2$,
3) the inequality gap, defined as 
$|| \boldsymbol{z}_T - \left[ \boldsymbol{K}(\boldsymbol{X}_T, \boldsymbol{X}_T; \boldsymbol{\theta}_h) + \sigma_{e}^{2} \boldsymbol{I}_{n} \right]^{-1}\boldsymbol{y}_{T}||_2^2$, and 
4) the ADMM objective value, $L_{\rho}(\boldsymbol{\theta}^{*}, \boldsymbol{z}_{T}^{*}, \boldsymbol{\lambda}^{*})$, obtained in one particular Monte-Carlo trial over a synthetic data set with the SE, LP, and SE+LP kernel respectively. In the plots, $x$-axis indicates the number of iterations and $y$-axis gives the corresponding values of the four measures.}
\label{fig:figure3}
\end{figure}

\paragraph{Convergence Property:}
GPCV-ADMM has its merit with rapid convergence to a moderately good estimation within a few iterations. The convergence curves of the selected parameters as well as the ADMM objective value are shown in Figure~\ref{fig:figure3} for one particular Monte Carlo trial. It is noticed that the number of iterations needed to converge is not influenced as the data size increases. In our experiments, we manually set the maximum number of iterations to be 100 such that both the CT and the risk of over-fitting could be well reduced.
\paragraph{Computational Complexity:}
We aim to verify that GPCV-ADMM has only $O(n^2)$ complexity while GPML has $O(n^3)$ complexity. Since the total number of iterations needed is not determined by increased sample size for both schemes, we only need to verify the complexity for each iteration. To this end, we fixed the kernel configuration and varied the data size from $n=500$ to 1000 and 2000, to see whether a quadratic increase in the computational time (CT) would be witnessed. Here, we treat the data size $n=500$ as the baseline, and compute the scaling factor (defined as $\frac{CT_{A}/CT_{B}}{n_{A}/n_{B}}$, where the subscript $B$ stands for the baseline data sets with $n=500$ and $A$ stands for data sets with larger sample sizes $n_A=1000, 2000$). 

\begin{figure}[H]
\centering
\includegraphics[width=\linewidth]{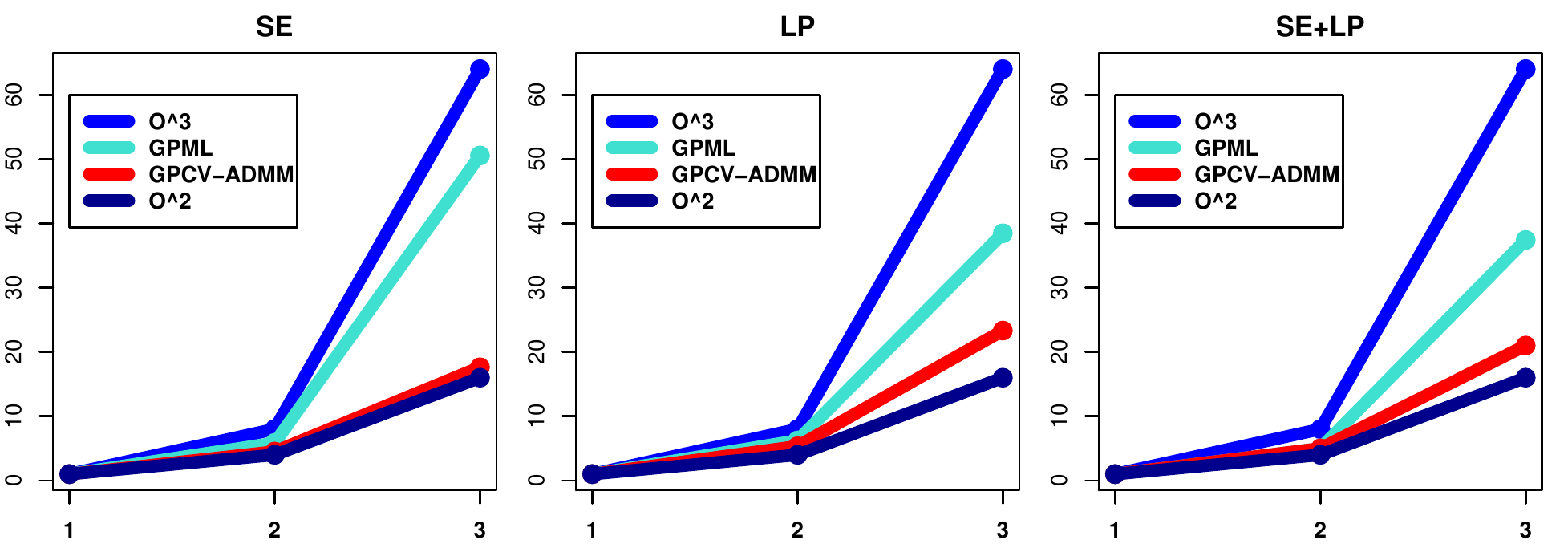} \\
\caption{Scaling factors of GPML and GPCV-ADMM on the synthetic data sets. In the plots, $x$-axis represents the ratio, $n_A/n_B$, and $y$-axis represents the corresponding scaling factor. The scaling factors corresponding $O(n^2)$ and $O(n^3)$ complexity generally are also drawn as references.} 
\label{fig:figure2}
\end{figure}

\begin{figure}[]
\centering
\includegraphics[width=0.9\linewidth]{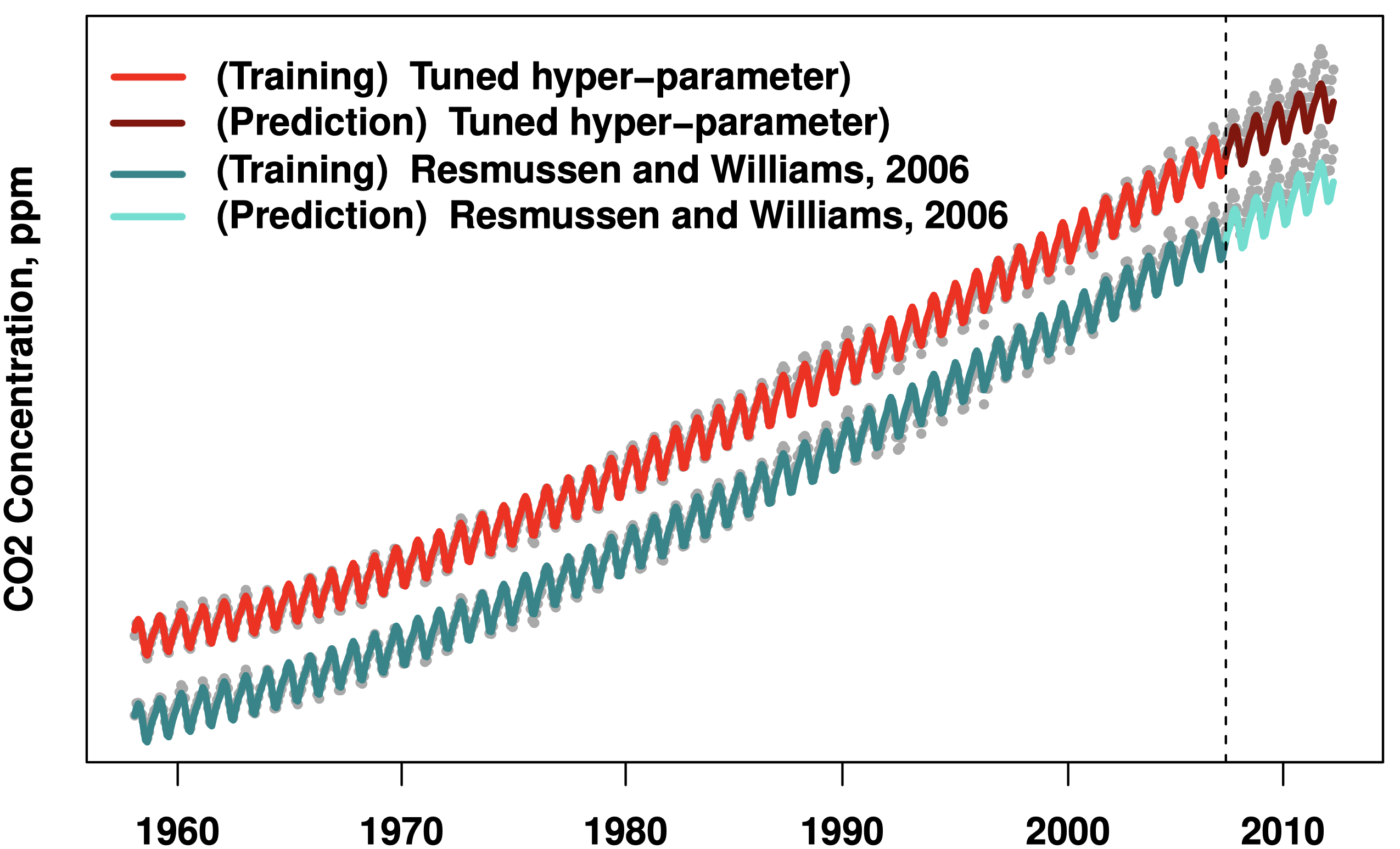} \\
\caption{Fitting and prediction performance of GPML and GPCV-ADMM on the $CO_2$ concentration data, with the real data points represented by gray dots. Since the two fitted lines are very close to each other, we slightly shift one curve for better visualization.}
\label{fig:figure4}
\end{figure}

The scaling factor across data sets generated from three kernels are shown in Figure~\ref{fig:figure2}. It is clear that the scaling factor of GPML is consistently larger than that of GPCV-ADMM. It is also noticed that a quadratic increase in the CT of GPCV-ADMM and a cubic increase in the CT of GPML would become more apparent as data size increases.

\subsection{Implementation Details}
The practical implementation of GPCV-ADMM requires special attentions to the following aspects.

\paragraph{Initialization:} A good starting point for both the hyper-parameters $\boldsymbol{\theta}$ and the auxiliary variable $\boldsymbol{z}_T$, will lead to faster and smoother convergence of GPCV-ADMM as observed in Figure~\ref{fig:figure3}. Random restarts could be adopted to alleviate the adverse impact of bad initializations. 

\paragraph{Numerical Search:} We follow (\ref{eq:gradient-descent for theta}) to update the GP hyper-parameters numerically. Coordinate descent \cite{Bertsekas16} is adopted when $\boldsymbol{\theta}$ has more than one element. New GD type of methods such as the ADAM algorithm \cite{Kingma15} and other variants could be used for faster and more stable numerical search. 

\paragraph{Choice of the regularization parameter $\rho$:} The magnitude of $\rho$ controls both the descent speed and the convexity of the ADMM objective function. A large $\rho$ endows a strong convexity of the ADMM objective function, yet often requiring more iterations to converge. A smaller $\rho$ endows faster descent, but the training procedure may get stuck at a bad local minimum more easily. When a suitable $\rho$ value is difficult to determine, one possible remedy, as suggested in \cite{Hong16}, is to use a different and smaller $\rho'$ in (\ref{eq:lambdaUpd}) for updating the dual variable.

\section{Conclusion}
\label{sec:Conclusion}

In this paper, we proposed two general CV based GP hyper-parameter optimization schemes suitable for big data applications. By introducing a nonlinear equality constraint to avoid large-scale matrix inverse, the resulting GPCV-ADMM scheme was proven to reduce the $O(n^3)$ computational complexity of the state-of-the-art GPML scheme considerably to $O(n^2)$. Unlike the existing low-complexity GP methods, GPCV-ADMM does not make any sophisticated approximations, and it reduces the gap between the training- and test performance, and most favorably, is extremely easy to implement. Especially, the $K$-fold CV based scheme has the potential to exploit the multi-core processing in modern computing platforms, and is robust to over-fitting. Various experimental results validated the performance of the proposed scheme, which in some cases outperforms its GPML counterpart with smaller sample standard deviation of the hyper-parameter estimates, lower test MSE, and most importantly, a significant reduction in the computational complexity. 

\newpage 
\section{Appendix (supplementary)}
\label{sec:Appendix}

\subsection{Partial Derivatives and Gradient}
The expression of $\boldsymbol{g}^{(\eta)}\left(\boldsymbol{\theta} \right)$ used for updating the GP hyper-parameters, $\boldsymbol{\theta}$, in (12) is obtained as:
\begin{align}
\boldsymbol{g}^{(\eta)}\left(\boldsymbol{\theta} \right) &= -2 \cdot \boldsymbol{y}_{V}^{T}  \boldsymbol{K}_{VT}(\boldsymbol{\theta}_h) \boldsymbol{z}_T^{\eta}  \nonumber \\
&+ (\boldsymbol{z}_T^{\eta})^{T} \boldsymbol{K}_{VT}(\boldsymbol{\theta}_h)^{T} \boldsymbol{K}_{VT}(\boldsymbol{\theta}_h) \boldsymbol{z}_T^{\eta} \nonumber \\
&+ \boldsymbol{\lambda}^{T} \boldsymbol{K}_{TT}(\boldsymbol{\theta}_h)  \boldsymbol{z}_T^{\eta}  \nonumber \\
&+ \rho(\sigma_{e}^{2}\boldsymbol{z}_T^{\eta} - \boldsymbol{y}_{T})^{T} \boldsymbol{K}_{TT}(\boldsymbol{\theta}_h)  \boldsymbol{z}_T^{\eta} \nonumber \\
&+ \frac{\rho}{2} (\boldsymbol{z}_T^{\eta})^{T} \boldsymbol{K}_{TT}(\boldsymbol{\theta}_h) \boldsymbol{K}_{TT}(\boldsymbol{\theta}_h) \boldsymbol{z}_T^{\eta}.
\end{align}
For each element of $\boldsymbol{\theta}$ (denoted as $\theta_i$), its partial derivative is computed as:
\begin{align}
\frac{\partial \boldsymbol{g}^{(\eta)}\left(\boldsymbol{\theta} \right)}{\partial \theta_i} &= -2 \cdot \boldsymbol{y}_{V}^{T}  \frac{\partial \boldsymbol{K}_{VT}(\boldsymbol{\theta}_h)}{\partial \theta_i}  \boldsymbol{z}_T^{\eta}  \nonumber \\
&+ (\boldsymbol{z}_T^{\eta})^{T} \frac{\partial \boldsymbol{K}_{VT}^{T}(\boldsymbol{\theta}_h)}{\partial \theta_i}   \boldsymbol{K}_{VT}(\boldsymbol{\theta}_h) \boldsymbol{z}_T^{\eta} \nonumber \\
&+ (\boldsymbol{z}_T^{\eta})^{T} \boldsymbol{K}_{VT}^{T}(\boldsymbol{\theta}_h) \frac{\partial \boldsymbol{K}_{VT}(\boldsymbol{\theta}_h)}{\partial \theta_i}   \boldsymbol{z}_T^{\eta} \nonumber \\
&+ \boldsymbol{\lambda}^{T} \frac{\partial \boldsymbol{K}_{TT}(\boldsymbol{\theta}_h)}{\partial \theta_i} \boldsymbol{z}_T^{\eta}  \nonumber \\
&+ \rho(\sigma_{e}^{2}\boldsymbol{z}_T^{\eta} - \boldsymbol{y}_{T})^{T} \frac{\partial \boldsymbol{K}_{TT}(\boldsymbol{\theta}_h)}{\partial \theta_i} \boldsymbol{z}_T^{\eta} \nonumber \\
&+ \frac{\rho}{2} (\boldsymbol{z}_T^{\eta})^{T} \boldsymbol{K}_{TT}(\boldsymbol{\theta}_h) \frac{\partial \boldsymbol{K}_{TT}(\boldsymbol{\theta}_h)}{\partial \theta_i}   \boldsymbol{z}_T^{\eta} \nonumber \\
&+ \frac{\rho}{2} (\boldsymbol{z}_T^{\eta})^{T} \frac{\partial \boldsymbol{K}_{TT}(\boldsymbol{\theta}_h)}{\partial \theta_i} \boldsymbol{K}_{TT}(\boldsymbol{\theta}_h)  \boldsymbol{z}_T^{\eta}.
\end{align}

\subsection{Explicit Form of Kernel Functions}

The expressions for the selected kernels that we use for the synthetic data are listed below. 

\begin{itemize}
\nonumber
\item  \textbf{Squared Exponential (SE) Kernel} \\
SE kernel is usually regarded as the default kernel for GP models, due to its great universality as well as many good properties. The length scale $l$ in an SE kernel specifies the width of the kernel and thereby determines the smoothness of the regression function. 
\begin{align}
k_{se}(x,x') = \sigma^2 exp \left( -\frac{(x-x')^2}{2l^2} \right)
\end{align}

\item \textbf{Locally Periodic (LP) Kernel}\\
Periodicity is another important pattern that people always get interested, especially in modeling time series data. Most of the real data do not repeat themselves exactly. Therefore combining a local kernel together with a periodic kernel into a locally periodic kernel, is considered to allow the shape of the repeating patterns to evolve over time:
\begin{align}
k&_{lp}(x,x') = \nonumber \\
&\sigma^2 exp \left( -\frac{2sin^2(\pi|x-x'|/p)}{l^2}\right) exp \left(-\frac{(x-x')^2}{2l^2} \right)
\end{align}

\item \textbf{Composite SE + LP Kernel}\\
One good thing about using kernel function is its flexibility in combining various kernel components, which allows multiplications and/or additions over different kernels to capture different features of the data. In our experiments, we added up one SE kernel and one LP kernel to model local periodicity with trend.
\begin{align}
k&_{se+lp}(x,x') = \sigma^2 exp \left( -\frac{(x-x')^2}{2l_1^2} \right)+ \nonumber \\ &\sigma^2 exp \left(-\frac{2sin^2(\pi|x-x'|/p)}{l_2^2}\right) exp\left(-\frac{(x-x')^2}{2l_2^2}\right)
\end{align}
\end{itemize}

%
\newpage
\bibliographystyle{ijcai19}
\bibliography{ijcai19}

\begin{thebibliography}{}

\bibitem[\protect\citeauthoryear{Ambikasaran \bgroup \em et al.\egroup
  }{2016}]{Ambikasaran16}
S.~Ambikasaran, D.~Foreman-Mackey, L.~Greengard, D.~W. Hogg, and M.~O'Neil.
\newblock Fast direct methods for {G}aussian processes.
\newblock {\em IEEE Trans. Pattern Recognition and Machine Intelligence},
  38(2):252--265, February 2016.

\bibitem[\protect\citeauthoryear{Bertsekas and Tsitsiklis}{1997}]{Bertsekas97}
D.~P. Bertsekas and J.~N. Tsitsiklis.
\newblock {\em Parallel and Distributed Computation: Numerical Methods, 2nd
  Edition}.
\newblock Athena Scientific, Belmont, MA. US., 1997.

\bibitem[\protect\citeauthoryear{Bertsekas}{2016}]{Bertsekas16}
D.~P. Bertsekas.
\newblock {\em Nonlinear Programming, 3rd. Edition}.
\newblock Athena Scientific, Belmont, MA. US., 2016.

\bibitem[\protect\citeauthoryear{Boyd \bgroup \em et al.\egroup
  }{2011}]{Boyd11}
S.~Boyd, N.~Parikh, E.~Chu, B.~Peleato, and J.~Eckstein.
\newblock Distributed optimization and statistical learning via the alternating
  direction method of multipliers.
\newblock {\em Foundation Trends Machine Learning}, 3(1):1--122, January 2011.

\bibitem[\protect\citeauthoryear{Deisenroth and Ng}{2015}]{Deisenroth15}
M.~P. Deisenroth and J.~W. Ng.
\newblock Distributed {Gaussian} processes.
\newblock In {\em Proc. of International Conference on Machine Learning
  (ICML)}, pages 1481--1490, Lille, France, July 2015.

\bibitem[\protect\citeauthoryear{Eckstein and Bertsekas}{1992}]{Eckstein92}
J.~Eckstein and D.~P. Bertsekas.
\newblock On the {Douglas---Rachford} splitting method and the proximal point
  algorithm for maximal monotone operators.
\newblock {\em Mathematical Programming}, 55(1):293--318, April 1992.

\bibitem[\protect\citeauthoryear{Frigola and Rasmussen}{2013}]{Frigola13}
R.~Frigola and C.~E. Rasmussen.
\newblock Integrated pre-processing for {Bayesian} nonlinear system
  identification with {Gaussian} processes.
\newblock In {\em Proc. of IEEE Conference on Decision and Control (CDC)},
  pages 5371--5376, Florence, Italy, December 2013.

\bibitem[\protect\citeauthoryear{Frigola \bgroup \em et al.\egroup
  }{2014}]{Frigola14}
R.~Frigola, Y.~Chen, and C.~E. Rasmussen.
\newblock Variational {Gaussian} process state-space models.
\newblock In {\em Proc. of Advances in Neural Information Processing Systems
  (NIPS)}, pages 3680--3688, Cambridge, MA, USA, 2014.

\bibitem[\protect\citeauthoryear{Han \bgroup \em et al.\egroup }{2016}]{Han16}
J.~Han, X.~Zhang, and F.~Wang.
\newblock {Gaussian} process regression stochastic volatility model for
  financial time series.
\newblock {\em IEEE Journal of Selected Topics in Signal Processing},
  10(6):1015--1028, September 2016.

\bibitem[\protect\citeauthoryear{Havasi \bgroup \em et al.\egroup
  }{2018}]{Havasi18}
M.~Havasi, J.M. Hernandez-Lobato, and J.~J. Murillo-Fuentes.
\newblock Inference in deep {G}aussian processes using stochastic gradient
  {Hamiltonian} {M}onte {C}arlo.
\newblock In {\em Proc. International Conference on Neural Information
  Processing Systems (NIPS)}, pages 1--10, Montreal, Canada, 2018.

\bibitem[\protect\citeauthoryear{Hensman \bgroup \em et al.\egroup
  }{2015}]{Hensman15}
J.~Hensman, A.~G. Matthews, M.~Filippone, and Z.~Ghahramani.
\newblock {MCMC} for variationally sparse {Gaussian} processes.
\newblock In {\em Proc. of Advances in Neural Information Processing Systems
  (NIPS)}, pages 1648--1656, Montreal, Canada, 2015.

\bibitem[\protect\citeauthoryear{Hong \bgroup \em et al.\egroup
  }{2016}]{Hong16}
M.~Hong, Z.~Q. Luo, and M.~Razaviyayn.
\newblock Convergence analysis of alternating direction method of multipliers
  for a family of nonconvex problems.
\newblock {\em SIAM Journal on Optimization}, 26(1):337--364, January 2016.

\bibitem[\protect\citeauthoryear{Kingma and Ba}{2015}]{Kingma15}
D.~P. Kingma and J.~Ba.
\newblock Adam: A method for stochastic optimization.
\newblock In {\em Proc. of International Conference on Learning Representations
  (ICLR)}, San Diego, US, 2015.

\bibitem[\protect\citeauthoryear{Krauth \bgroup \em et al.\egroup
  }{2017}]{Krauth17}
K.~Krauth, E.V. Bonilla, Cutajar K., and Filippone M.
\newblock {AutoGP}: Exploring the capabilities and limitations of {Gaussian}
  process models.
\newblock In {\em Proc. of Conference on Uncertainty in Artificial Intelligence
  (UAI)}, Sydney, Australia, August 2017.

\bibitem[\protect\citeauthoryear{Liu \bgroup \em et al.\egroup }{2018}]{Liu18b}
H.~Liu, Y.-S. Ong, X.~Shen, and J.~Cai.
\newblock When {Gaussian} process meets big data: A review of scalable {GPs}.
\newblock {\em https://arxiv.org/abs/1807.01065}, 2018.

\bibitem[\protect\citeauthoryear{MacKay}{1998}]{MacKay98}
D.~J.~C. MacKay.
\newblock An introduction to {Gaussian} processes, 1998.

\bibitem[\protect\citeauthoryear{Matthews \bgroup \em et al.\egroup
  }{2018}]{Mattews18}
A.~G. Matthews, J.~Hron, M.~Rowland, R.~E. Turner, and Z.~Ghahramani.
\newblock {Gaussian} process behaviour in wide deep neural networks.
\newblock In {\em Proc. of International Conference on Learning Representations
  (ICLR)}, Vancouver, Canada, 2018.

\bibitem[\protect\citeauthoryear{Neal}{1996}]{Neal96}
R.~M. Neal.
\newblock {\em Bayesian Learning for Neural Networks: Lecture Notes in
  Statistics}.
\newblock Springer, New York, US., 1996.

\bibitem[\protect\citeauthoryear{Neal}{1997}]{Neal97}
R.~M. Neal.
\newblock Monte {Carlo} implementation of {Gaussian} process models for
  {Bayesian} regression and classification.
\newblock Technical Report Technical Report No. 9702, Dept. of Statistics,
  University of Toronto, 1997.

\bibitem[\protect\citeauthoryear{Qui\~{n}onero Candela and
  Rasmussen}{2005}]{Candela05}
J.~Qui\~{n}onero Candela and C.~E. Rasmussen.
\newblock A unifying view of sparse approximate {G}aussian process regression.
\newblock {\em Journal Machine Learning Research}, 6(1):1939--1959, December
  2005.

\bibitem[\protect\citeauthoryear{Rasmussen and Williams}{2006}]{RW06}
C.~E. Rasmussen and C.~I.~K. Williams.
\newblock {\em Gaussian Processes for Machine Learning}.
\newblock MIT Press, 2006.

\bibitem[\protect\citeauthoryear{Titsias}{2009}]{Titsias09}
M.~K. Titsias.
\newblock Variational learning of inducing variables in sparse {G}aussian
  processes.
\newblock In {\em Proc. of International Conference on Artificial Intelligence
  and Statistics (AISTATS)}, pages 567--574, Clearwater Beach, Florida, USA,
  April 2009.

\bibitem[\protect\citeauthoryear{Wilson and Nickisch}{2015}]{Wilson15}
A.~G. Wilson and H.~Nickisch.
\newblock Kernel interpolation for scalable structured {Gaussian} processes
  ({KISS-GP}).
\newblock In {\em Proc. of International Conference on Machine Learning
  (ICML)}, pages 1775--1784, Lille, France, 2015.

\end{thebibliography}

\end{document}